%% file: root.tex
\documentclass[letterpaper, 10 pt, conference]{ieeeconf}  

\IEEEoverridecommandlockouts                              

\usepackage{graphics} 
\usepackage{epsfig} 
\usepackage{amsmath} 
\usepackage{amssymb}  
\usepackage{url}
\usepackage{booktabs}
\usepackage[dvipsnames]{xcolor}
\usepackage{multirow}
\usepackage{placeins}

\usepackage{cite}
\usepackage[bookmarks=false]{hyperref}

\title{\LARGE \bf
BEVPose: Unveiling Scene Semantics through Pose-Guided Multi-Modal BEV Alignment}

\author{Mehdi Hosseinzadeh$^{1}$ and Ian Reid$^{1,2}$
\thanks{$^{1}$Australian Institute for Machine Learning, The University of Adelaide, Australia.
       }%
\thanks{$^{2}$Mohamed Bin Zayed University of Artificial Intelligence, UAE.
        }%
}

\begin{document}

\maketitle
\thispagestyle{empty}
\pagestyle{empty}


\input{sec/0_abstract}

\input{sec/1_intro}

\input{sec/2_related_work}

\input{sec/3_bev_representation}
\input{sec/4_bev_alignment}
\input{sec/5_experiments}

\input{sec/6_conclusion}





\bibliographystyle{IEEEtran}
\bibliography{references}

\end{document}

%% file: sec/0_abstract.tex
\begin{abstract}

In the field of autonomous driving and mobile robotics, there has been a significant shift in the methods used to create Bird's Eye View (BEV) representations. This shift is characterised by using transformers and learning to fuse measurements from disparate vision sensors, mainly lidar and cameras, into a 2D planar ground-based representation. However, these learning-based methods for creating such maps often rely heavily on extensive annotated data, presenting notable challenges, particularly in diverse or non-urban environments where large-scale datasets are scarce. In this work, we present BEVPose, a framework that integrates BEV representations from camera and lidar data, using sensor pose as a guiding supervisory signal. This method notably reduces the dependence on costly annotated data. By leveraging pose information, we align and fuse multi-modal sensory inputs, facilitating the learning of latent BEV embeddings that capture both geometric and semantic aspects of the environment. Our pretraining approach demonstrates promising performance in BEV map segmentation tasks, outperforming fully-supervised state-of-the-art methods, while necessitating only a minimal amount of annotated data. This development not only confronts the challenge of data efficiency in BEV representation learning but also broadens the potential for such techniques in a variety of domains, including off-road and indoor environments. 

\end{abstract}

%% file: sec/1_intro.tex
\section{Introduction}
\label{sec:intro}

Mapping and reconstructing the environment is pivotal in the domain of autonomy, be it for mobile robotics or autonomous vehicles. Understanding the surroundings is not merely a precursor but a necessity for subsequent tasks such as navigation, planning or manipulation within these environments. Decades of research across robotics and computer vision have been dedicated to this endeavour, adopting diverse methods, representations, and sensors tailored to specific tasks. This ranges from online sequential methods like Simultaneous Localisation and Mapping (SLAM) \cite{ORBSLAM3_TRO, orbslam, lsd-slam} to more batch-type solutions in Structure-from-Motion (SfM) \cite{schoenberger2016mvs, schoenberger2016sfm}, utilising various sensor modalities, from cameras \cite{orbslam, engel2016direct} to lidars \cite{legoloam2018, wang2021}. Representations vary from sparse landmarks with either hand-crafted \cite{lsd-slam, orbslam} or deep-learned features \cite{bloesch2019codeslam} to dense 3D forms such as meshes and voxels, including implicit formats from classic signed distance functions to neural representations \cite{bylow2013real,wang2021nerf,pumarola2021d}. The choice of representation - geometric landmarks or semantically meaningful entities \cite{orbslam, yang2019cubeslam, taguchi2013point, hosseinzadeh2019real, hosseinzadeh2019structure},  - further diversifies mapping approaches.

However, no single ``universally optimal'' solution exists that encapsulates all these facets to deliver a general mapping solution. The right representation, sensor inputs, and the desired balance between geometric and semantic information are crucial in determining the mapping approach.

The evolution of Bird's Eye View (BEV) representations in recent years has been significantly influenced by the growing needs of the autonomous driving industry, where accurately determining the 3DOF pose of objects and semantics in a planar world plays an important role. This is particularly crucial in applications such as 2D/3D object detection \cite{huang2021bevdet, huang2022bevdet4d} and 2D map semantic segmentation \cite{bevfusion2022, liang2022bevfusion}.
BEV maps provide a unified representation, merging data from various sensors into a single, consistent format. This integration, in line with the planar geometry of the scene, enhances the learning of rich representations essential for 2D projections of the world, required in specific applications.

One major challenge in employing this approach, akin to other fully-supervised methods, is the extensive annotation required. For BEV-related tasks, annotations must be provided from a different perspective than the input sensor modality, imposing additional costs and constraints. This is especially problematic when transferring this knowledge to different scenarios, such as non-urban or indoor environments, where large-scale, high-quality datasets are scarce.

The self-supervised paradigm, gaining traction in recent years, offers a solution, albeit underexplored in the context of self-supervised BEV representation \cite{sarlin2023orienternet, sarlin2023snap} for multi-modal fusion, specifically with cameras and lidars. Our work, BEVPose, addresses this gap. We leverage the relatively inexpensive and readily available ground-truth data, i.e. sensor poses, to learn and fuse BEV representations from camera and lidar inputs. The known relative sensor poses provide a rich supervisory signal for learning latent BEV embeddings and fusing modalities by aligning BEV maps. Distinct map features and landmarks emerge implicitly in the learned ``self-pose-supervised'' BEV representations, serving as a solid foundation for subsequent fine-tuning for downstream tasks like map semantic segmentation. Our results demonstrate that we can outperform state-of-the-art BEV map segmentation methods, using only a fraction of the costly annotated data they require.

In summary, we present BEVPose, a method founded on self-supervised, multi-modal pretraining with pose supervision at its core. Our approach aims to enhance the accuracy of lifting 2D perspective camera features into 3D space, while reducing the reliance on annotated data for attention-based multi-modal fusion. By aligning neural BEV maps from cameras and lidar, and simultaneously learning implicit depth distributions, BEVPose not only facilitates effective pretraining for downstream tasks like map semantic segmentation but also outperforms existing fully-supervised methods. Furthermore, our method distinguishes itself through its data efficiency, demonstrating the ability to match the performance levels of fully-supervised approaches with significantly less annotated data, and achieving comparable outcomes even when limited to just a fifth of the data for fine-tuning, with only minimal compromise.

%% file: sec/2_related_work.tex
\section{Related Work}
\label{sec:related}

While we have already mentioned some related works in Section~\ref{sec:intro}, we summarise them here categorically for ease of referencing and to consolidate related research.

\textbf{BEV Perception:}
Recent advances in BEV map segmentation have seen a variety of innovative approaches. BEVFusion \cite{bevfusion2022, liang2022bevfusion} integrate multi-sensor data into a unified BEV representation \cite{bevfusion2022}. Similarly, the works of \cite{yin2020center, huang2023dal, huang2022bevpoolv2, huang2022bevdet4d, huang2021bevdet}, FCOS3D \cite{tian2019fcos}, and DETR3D \cite{carion2020end} integrate camera and LiDAR data for enhanced 3D perception. The fusion of multi-modal sensory data is further explored in papers such as \cite{chen2017multi}, \cite{qi2018frustum}, and \cite{wang2021end}, each proposing methods for object detection and BEV map segmentation. Additionally, works like \cite{reading2021categorical} and LSS \cite{philion2020lift} have made progress in camera-based 3D perception, converting camera features to BEV and extending existing models for 3D object detection. 
These methods can be categorised into two groups: lifting-based approaches and transformer/attention-based approaches. In the former, 2D perspective-view features are projected into 3D space using various mechanisms, ranging from simple parameter-free unprojection via bilinear sampling, to explicit depth estimation employing off-the-shelf depth prediction methods, and even to implicit depth learning techniques \cite{philion2020lift, you2019pseudo, li2023voxformer}.
In transformer attention-based approaches, self-attention and cross-attention operations are utilised to integrate multi-modal features into a predefined BEV grid \cite{saha2022translating, zhou2022cross, li2022bevformer, peng2023bevsegformer}. For instance, in transformer terminology, BEV embeddings from one modality act as queries, while other modalities serve as keys and values. Although these methods typically yield more accurate results, they require significantly higher computational resources, especially for higher-resolution grids.
The integration of temporal cues in multi-camera 3D object detection, as studied in \cite{zhu2020deformable, liu2022petr, liu2022petrv2, zhang2022beverse}, represents another direction. These efforts collectively contribute to the ongoing evolution of BEV map segmentation, demonstrating the potential of multi-modal fusion and advanced perception techniques in this field.

\textbf{Self-Supervised Learning:}
Self-supervised learning has recently gained momentum, with numerous studies leveraging unlabeled datasets for representation learning useful in downstream tasks. Key contributions in this field include \cite{vandenoord2018representation}, \cite{he2020momentum}, and work on self-supervised vision transformers \cite{caron2021emerging}. \cite{pinheiro2020unsupervised} and \cite{he2022masked} have made progress in dense visual representations and scalable vision learners, respectively. \cite{larsson2019fine} have explored fine-grained segmentation networks, while NeRF \cite{mildenhall2020nerf} represents scenes as neural radiance fields for view synthesis. Recent advancements also include the work on static-dynamic disentanglement \cite{sharma2022seeing} and CoCoNets \cite{lal2021coconets} for continuous contrastive 3D scene representations.

%% file: sec/3_bev_representation.tex
\section{BEV Representation}
\label{sec:bev}

In our approach, we utilise the BEV map as a compact yet comprehensive method for reconstructing the surrounding 3D scene. 
It provides a unified framework for integrating multi-modal sensory data and embedding semantic layers, thereby enriching the overall scene comprehension. 

\begin{figure*}
  \centering
  \includegraphics[width=\linewidth]{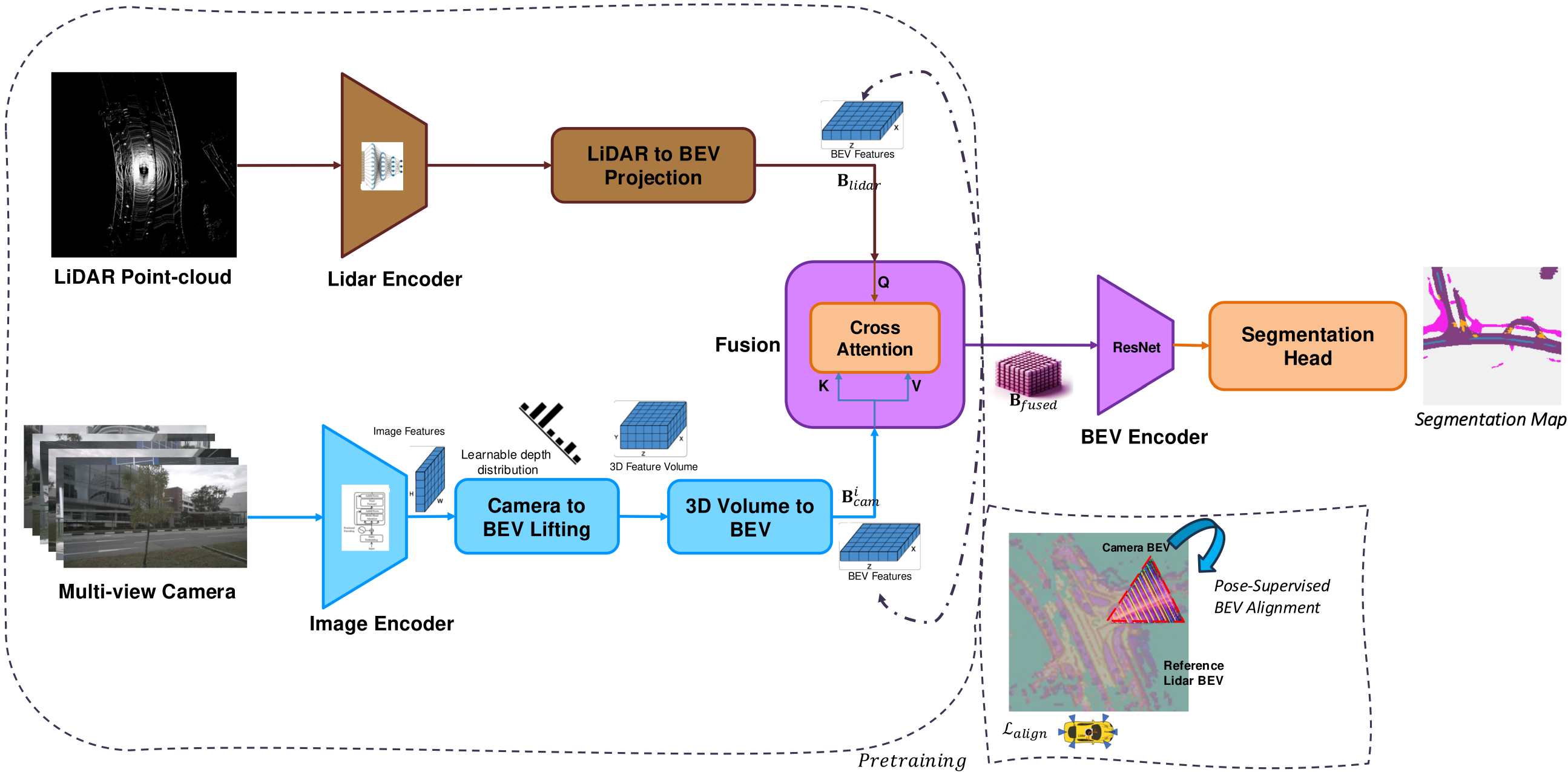}
  \caption{\textbf{BEVPose Framework.} This figure illustrates our framework for learning latent BEV representations through alignment, using pose as a supervisory signal within a contrastive learning paradigm. Multi-view camera images and lidar point-cloud data are encoded, transformed, and fused into a latent BEV space. By aligning BEV features in a self-supervised manner, the resulting BEV representation captures high-level semantic and geometric characteristics of the scene. These features are then fine-tuned for downstream tasks such as semantic segmentation, requiring significantly less annotated data compared to fully supervised BEV segmentation methods.}
  \label{fig:framework}
\end{figure*}

We utilise two distinct and complementary sensory input modalities in our model: a multi-view camera setup consisting of \(N_c\) calibrated surrounding cameras, and a full-circle \(360^\circ\) lidar point-cloud. We presuppose that both the cameras and the lidar are calibrated, i.e., camera intrinsics (\(K_{c_i} \in \mathbb{R}^{3\times3}\)) and extrinsic transformations between cameras and lidar (\(T^{c_i}_{c_j}, T^{l}_{c_i} \in \mathrm{SE(3)}\)) are known. The geometry and semantics of the scene are encapsulated in a 2D grid BEV map, \(\mathbf{B}\), on the \(xz\) plane, centred at the lidar's origin. Each cell of \(\mathbf{B}\) corresponds to a pillar of a \(D\)-dimensional feature vector, e.g., \(\mathbf{B}_{{x_i}{z_j}} \in \mathbb{R}^D\), encoding the geometry and semantics pertinent to that specific grid cell.
An illustrative overview of our framework is provided in Figure~\ref{fig:framework}.

\subsection{Image Encoder}
Each RGB image (\(\mathbb{R}^{H \times W \times 3}\)) is encoded utilising a shared backbone and neck architecture into an image feature space of dimension \(C_c\). To transform these encoded features into the BEV space, the depth distribution over the features is requisite. 
Inspired by LSS~\cite{philion2020lift}, we lift the camera features into the BEV space by learning the discrete depth distribution of each image feature along its projection ray. After learning this distribution, the corresponding feature vectors are redistributed along the ray according to weights coming from the distribution. 
The depth distribution along the ray is treated as a finite probability mass function over \(L\) uniformly quantised depth values. The BEV pooling process, transforms the feature volume (\(\mathbb{R}^{H \times W \times L \times C_c}\)) of each camera into the BEV grid, by aggregating and flattening features in each grid cell along the \(y\)-axis (vertical direction) to produce a camera BEV map \(\mathbf{B}_{cam}^{i}\). Subsequently, all individual camera BEV maps are integrated to form the multi-view camera BEV map \(\mathbf{B}_{cam}\).
To expedite the pooling process, we leverage the pre-established camera calibrations, enabling the pre-calculation of associations between each pillar of the camera BEV volume and the final BEV grid map.

\subsection{LiDAR Encoder} 
The lidar point-cloud is initially voxelised using a predetermined resolution and subsequently input into the 3D encoder. 
This process yields a 4D feature volume (\(\mathbb{R}^{{L_x}\times{L_z}\times{L_y}\times{C_l}}\)). This feature volume is then vertically pooled along the \(y\)-axis to generate the lidar BEV map, denoted as \(\mathbf{B}_{lidar}\).

\subsection{Multi-modal Fusion} 
We obtain the final fused BEV map by utilising a cascade of four transformer blocks. In the cross-attention module, the lifted image features serve as keys and values, while the queries are derived from the lidar BEV features, inspired by \cite{bai2022transfusion}. This fused embedding is then passed through a bottleneck ResNet encoder \cite{he2016deep}, functioning as our BEV encoder, as illustrated in Figure~\ref{fig:framework}.


%% file: sec/4_bev_alignment.tex
\section{BEV Alignment}
\label{sec:bev_alignment}

As introduced in Section~\ref{sec:intro}, inspired by the successful studies in self-supervised learning \cite{vandenoord2018representation, lal2021coconets, sarlin2023snap}, we frame our BEV feature pretraining as a self-supervised BEV alignment problem, underpinned by pose supervision derived from the relative pose priors of calibrated multi-view cameras and lidar sensors. Our aim is to align these BEV maps to learn latent embeddings that are sufficiently rich to encapsulate both semantic and geometric nuances of the scene, which are imperative for the efficacious alignment of features.

Following established practices in self-supervised methods \cite{pinheiro2020unsupervised, carion2020end}, we formulate our alignment as a contrastive learning problem, by incorporating a contrastive loss, as outlined in Equation~\ref{eq:loss}, to embed the alignment objective within our framework. 
One approach to framing this alignment problem involves regarding the lidar BEV, \(\mathbf{B}_{lidar}\), as the reference BEV map, and each individual camera frustum BEV map, \(\mathbf{B}_{cam}^{i}\), as the corresponding query. These camera queries and lidar reference maps are defined within their respective local coordinate grids. The goal is to maximise the consistency between the query and reference BEV maps when using the ground-truth relative pose (positive sample), \({T^{l}_{c}}^{+}\), and to minimise it across all other incorrect relative poses (negative samples), \(T^{l}_{c_j}\). This objective translates to minimising the contrastive loss, articulated as:
\begin{equation}
  \mathcal{L}_{align} = -\log{\frac{\exp(\mathfrak{s}(\mathbf{B}_{cam}, \mathbf{B}_{lidar}; {T^{l}_{c}}^{+}) / \tau)}{\sum_{j=1}^{N_s} \exp(\mathfrak{s}(\mathbf{B}_{cam}, \mathbf{B}_{lidar}; T^{l}_{c_j}) / \tau)}}
  \label{eq:loss}
\end{equation}
where \(N_s\) represents the total number of samples and \(\mathfrak{s}(\mathbf{B}_{cam}, \mathbf{B}_{lidar}; T^{l}_{c})\) calculates the similarity score or the consistency between two BEV maps. This is achieved by summing the inner products of corresponding feature vectors after aligning and warping them into the same BEV grid map. 
Furthermore, \(\tau\) serves as a learnable temperature parameter within this framework.

To elucidate, consider \(p_c\) as a representative point within a specific camera BEV grid cell. This point is translated to the lidar frame using \(T^{l}_{c}\). Subsequently, the corresponding lidar BEV feature is deduced by interpolating the lidar BEV features of the relevant grid cell corners. The similarity measure is then computed as the sum of inner products of the correspondingly aligned BEV features, as elaborated in the following equation:
\begin{equation}
    \mathfrak{s}(\mathbf{B}_{cam}, \mathbf{B}_{lidar}; T^{l}_{c}) = \sum_{x,z} \hat{\mathbf{B}}_{{cam}_{xz}}^{\top}[x, z, :] \hat{\mathbf{B}}_{{lidar}_{xz}}[\mathfrak{i}(p_l), :]
  \label{eq:loss_similarity}
\end{equation}
Here, \(\hat{\mathbf{B}}[., ., :]\) represents the BEV feature in the respective grid, \(p_l\) is the transformation of \(p_c\), i.e. \(p_l = T^{l}_{c} p_c\), and \(\mathfrak{i}(.)\) denotes the interpolation function at a grid cell.

An extensively explored aspect of contrastive learning pertains to the selection of negative samples \cite{he2020momentum}. To pinpoint hard negatives—those with a high likelihood yet incorrect predictions, thereby guiding the probability distribution towards the ground truth—we utilise a strategy that samples negative poses from a distribution centred on the poses of other non-overlapping cameras.

%% file: sec/5_experiments.tex
\section{Experiments}
\label{sec:experiments}

We have undertaken a series of experiments to assess the performance of our pose-supervised pretraining approach, with a particular focus on its application to BEV map segmentation task. 
This task is chosen as a representative example to demonstrate the efficacy of our approach in capturing both the semantic and geometric aspects of the scene. 

During the ``self-pose-supervised'' pretraining phase, our model undergoes initial training across the full extent of the training dataset, deliberately excluding segmentation annotations. Following this preparatory stage, the model is further refined through fine-tuning, utilising a select fraction of the training set that includes (resource-intensive) annotations, with the segmentation head now integrated into the model.

\subsection{Experiments Setup and Architectures}
Our experiment setup adopts the widely recognised strategy of initially pretraining the model using an extensive corpus of unlabelled data, subsequently refining its performance through fine-tuning on a considerably smaller dataset of annotated examples for the specified downstream task.
Following \cite{huang2021bevdet, bevfusion2022}, our experimental setup employs the Swin-Transformer \cite{liu2021swin} as the image backbone, FPN \cite{lin2017feature} as the neck of the image encoder, and VoxelNet \cite{zhou2018voxelnet} as the lidar backbone. Camera images are resized to dimensions of \((H, W) = (256, 704)\), while lidar point-clouds are voxelised at a resolution of \(0.1\)m. The camera features, once lifted, are distributed over a depth range discretised into \([1, 80]\) meters with a step size of \(0.5\)m, resulting in \(L = 158\) discrete depth values. Optimisation processes utilise AdamW \cite{loshchilov2017decoupled} with a weight decay set to \(10^{-2}\). All experiments were conducted using four Nvidia A6000 GPUs.

We employ the nuScenes dataset \cite{caesar2020nuscenes} in our study, which is a widely-used, large-scale public outdoor urban dataset. It features 850 scenes for training and validation purposes, and an additional 150 scenes for testing. The dataset is equipped with data from \(N_c = 6\) cameras and \(1\) lidar, with the six cameras collectively achieving a \(360^\circ\) field-of-view. For our BEV map segmentation task, we focus on the annotations for 6 background classes (commonly referred to as 'stuff').

For the self-supervised pretraining, our model undergoes pretraining using the full training set, without employing segmentation annotations.
Following this initial phase, a select fraction of the training set (from 10\% to 50\%) - comprising data with (expensive) annotations - is employed to fine-tune our model, now with the segmentation head integrated, to refine its performance further.

\subsection{Efficacy of Pretraining and Segmentation Evaluations}

Following the self-supervised pretraining phase, we align with the practices in \cite{zhou2022cross} and \cite{bevfusion2022} to incorporate a task-specific head for semantic segmentation. We then compare the accuracy of our segmentation against baseline models and other state-of-the-art methods in this task. 
For classifying each semantic category within the specified ``stuff'' labels, we employ a binary semantic segmentation decoder. This decoder is equipped with three upsampling layers, each upscaling the latent BEV representation by a factor of 2, to its final dimension. To train this segmentation head, a focal loss \cite{lin2017focal} is utilised.

Our comparison adheres to the standard settings established after \cite{philion2020lift}, such as a map size and resolution defined as a \(100 \times 100\) meters region around the ego-car, with a resolution of \(0.5\)m per grid cell. The Intersection-over-Union (IoU) and mean IoU (mIoU) metrics are used for each semantic label as the primary indicator of model performance accuracy. 

We benchmark our framework's performance against a range of baselines and competitive approaches across both singular modalities (either camera or lidar) and multi-modal (camera+lidar) configurations. Our objective is to quantify the performance enhancement attributable to our self-supervised pretraining. Initially, we employ the entire training dataset for both pretraining and fine-tuning phases to compare our model's effectiveness against fully-supervised counterparts. The outcomes are summarised in Table~\ref{tab:evaluations}.

\begin{table*}
  \centering
  \caption{\textbf{Evaluation of BEV Map Segmentation on the nuScenes Dataset.} Our approach outperforms both single and multi-modal baselines, including the state-of-the-arts. Initiated with self-supervised pretraining using pose data, it is fine-tuned with ground-truth segmentation across the complete training set. The abbreviations 'C', 'L', and 'C+L' denote camera, lidar, and camera+lidar modalities, respectively. The mean IoU (mIoU) for each method is listed in the final column.}
  \begin{tabular}{@{}lcccccccc@{}}
    \toprule
    Method & Modality & carpark & walkway & lane & drivable & stop\_line & crossing &  mean \\
    \midrule
    LSS \cite{philion2020lift} & C & 39.1 & 46.3 & 36.5 & 75.4 & 30.3 & 38.8 &  44.4 \\
    CVT \cite{zhou2022cross} & C & 35.0 & 39.9 & 29.4 & 74.3 & 25.8 & 36.8 &  40.2 \\
    OFT \cite{roddick2018orthographic} & C & 35.9 & 45.9 & 33.9 & 74.0 & 27.5 & 35.3 &  42.1 \\
    M$^2$BEV \cite{xie2022m} & C & - & - & 40.5 & 77.2 & - & - &  - \\
    \midrule
    CenterPoint \cite{yin2021center} & L & 31.7 & 57.5 & 41.9 & 75.6 & 36.5 & 48.4 &  48.6 \\
    PointPillars \cite{lang2019pointpillars} & L & 27.7 & 53.1 & 37.5 & 72.0 & 29.7 & 43.14 &  43.8 \\
    \midrule
    \midrule
    MVP \cite{yin2021multimodal} & C+L & 33.0 & 57.0 & 42.2 & 76.1 & 36.9 & 48.7 &  49.0 \\
    PointPainting \cite{vora2020pointpainting} & C+L & 34.5 & 57.1 & 41.9 & 75.9 & 36.9 & 48.5 &  49.1 \\
    BEVFusion \cite{bevfusion2022} & C+L & 57.0 & 67.6 & 53.7 & 85.5 & 52.0 & 60.5 &  62.7 \\
    BEVPose (Ours) & C+L & \textbf{62.1} & \textbf{72.5} & \textbf{58.9} & \textbf{89.6} & \textbf{57.2} & \textbf{66.5} &  \textbf{67.8} \\ 
    \bottomrule
  \end{tabular}
  \label{tab:evaluations}
\end{table*}

Our method consistently surpasses both single and multi-modal benchmarks as well as the state-of-the-art (overall improvement of 5.1\%) across various categories, as demonstrated by the IoU and mIoU metrics in Table~\ref{tab:evaluations}. This underscores the significant performance uplift achieved through our self-supervised pretraining, which notably incurs a lower annotation cost compared to the extensive segmentation masks necessary for fine-tuning.

\begin{figure}[!t]
  \centering
  \includegraphics[width=\linewidth]{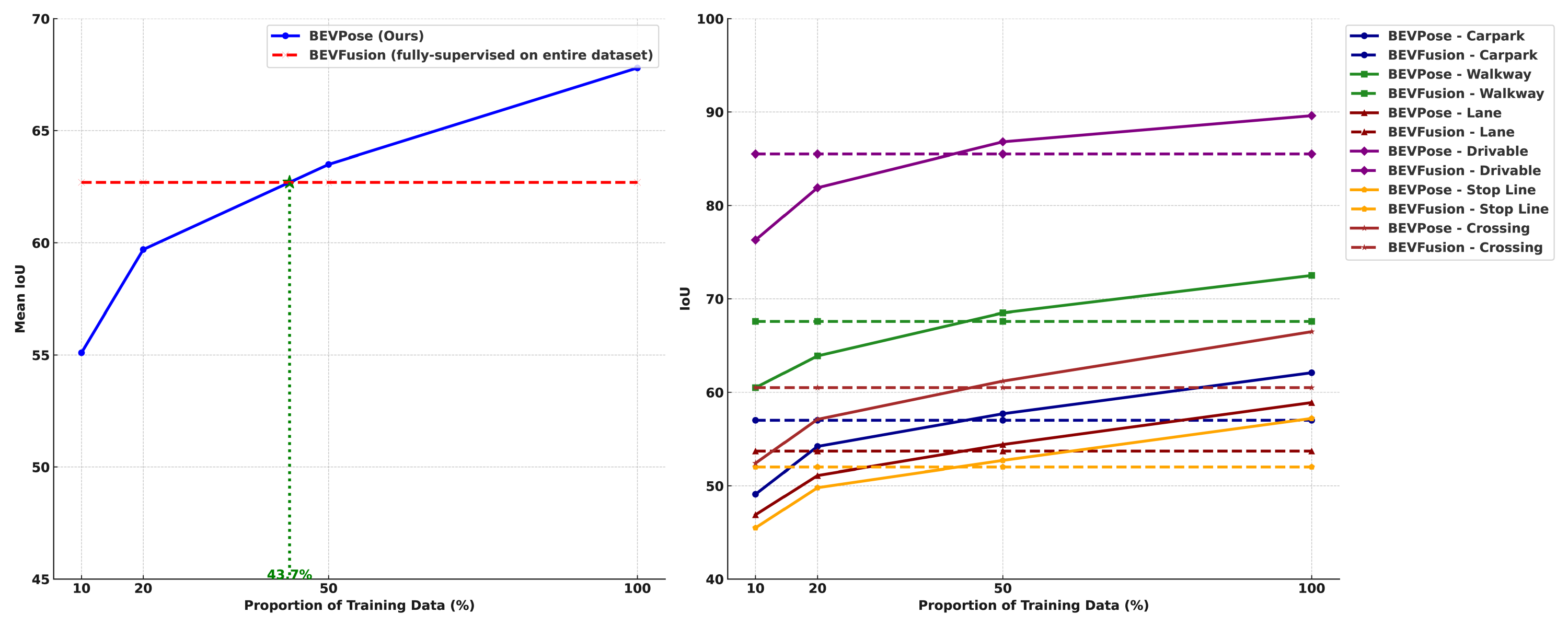}
  \caption{\textbf{BEVPose Performance Across Training Data Proportions.} This figure presents a detailed comparison of the BEVPose model, employing self-supervised pretraining, against the fully-supervised method of BEVFusion~\cite{bevfusion2022}, across a range of training dataset proportions used for fine-tuning. It is noteworthy that BEVFusion~\cite{bevfusion2022} utilises the full dataset for training. The left graph illustrates the mean IoU achieved by both approaches, underscoring that BEVPose equates to BEVFusion~\cite{bevfusion2022} with less than 44\% of the annotated training data, thus showcasing the value of self-supervised learning when leveraging a smaller quantity of ground-truth annotations. The right graph extends this analysis to the performance in individual categories (carpark, walkway, lane, drivable, stop line, and crossing). This comparison reveals BEVPose's potential for real-world deployment where the availability of annotated data might limit fully supervised methods.}
  \label{fig:bevpose_bevfusion_comparison}
\end{figure}

While the performance gains are crucial, another key consideration, especially relevant to autonomous driving and robotics, is data efficiency. In subsequent experiments, we evaluate our model's efficacy when fine-tuned on a limited portion of the training dataset, benchmarking against fully-supervised methods such as BEVFusion~\cite{bevfusion2022}. As illustrated in Table~\ref{tab:dataset-proportions} and Figure~\ref{fig:bevpose_bevfusion_comparison}, BEVPose matches \cite{bevfusion2022} with less than 44\% of the annotated training data and achieves comparable results with just 20\% of the data for fine-tuning. This data efficiency suggests a minor trade-off in overall performance may be acceptable in various applications, given the reduced annotation requirement. It also renders our approach particularly suitable for transfer to other domains, especially off-road rough terrain settings, where large-scale, high-quality datasets akin to nuScenes\cite{caesar2020nuscenes} for urban environments are not readily available.

\begin{table*}
  \centering
  \caption{\textbf{Efficacy of Self-Supervised Pretraining.} Evaluation results of BEVPose, leveraging self-supervised pretraining with pose supervision, against the fully-supervised benchmark method proposed in \cite{bevfusion2022}. This evaluation spans various proportions of the training dataset, all utilising ground-truth annotations for fine-tuning. 'C+L' denotes the use of both camera and lidar modalities. The final column represents the mean Intersection over Union (mIoU).}
  \begin{tabular}{c|l|cccccccc@{}}
    \toprule
    Training Data & Method & Modality & carpark & walkway & lane & drivable & stop\_line & crossing & mean \\
    \midrule
    100\% & BEVFusion \cite{bevfusion2022} & C+L & 57.0 & 67.6 & 53.7 & 85.5 & 52.0 & 60.5 &  62.7 \\
    \midrule
    10\% & BEVPose (Ours) & C+L & 49.1 & 60.5 & 46.9 & 76.3 & 45.5 & 52.4 &  55.1 \\ 
    \midrule
    20\% & BEVPose (Ours) & C+L & 54.2 & 63.9 & 51.1 & 81.9 & 49.8 & 57.1 &  59.7 \\ 
    \midrule
    50\% & BEVPose (Ours) & C+L & \textbf{57.7} & \textbf{68.5} & \textbf{54.4} & \textbf{86.8} & \textbf{52.7} & \textbf{61.2} &  \textbf{63.5} \\ 
    \bottomrule
  \end{tabular}
  \label{tab:dataset-proportions}
\end{table*}

\subsection{Qualitative Results}
\begin{figure*}
  \centering
  \includegraphics[width=\linewidth]{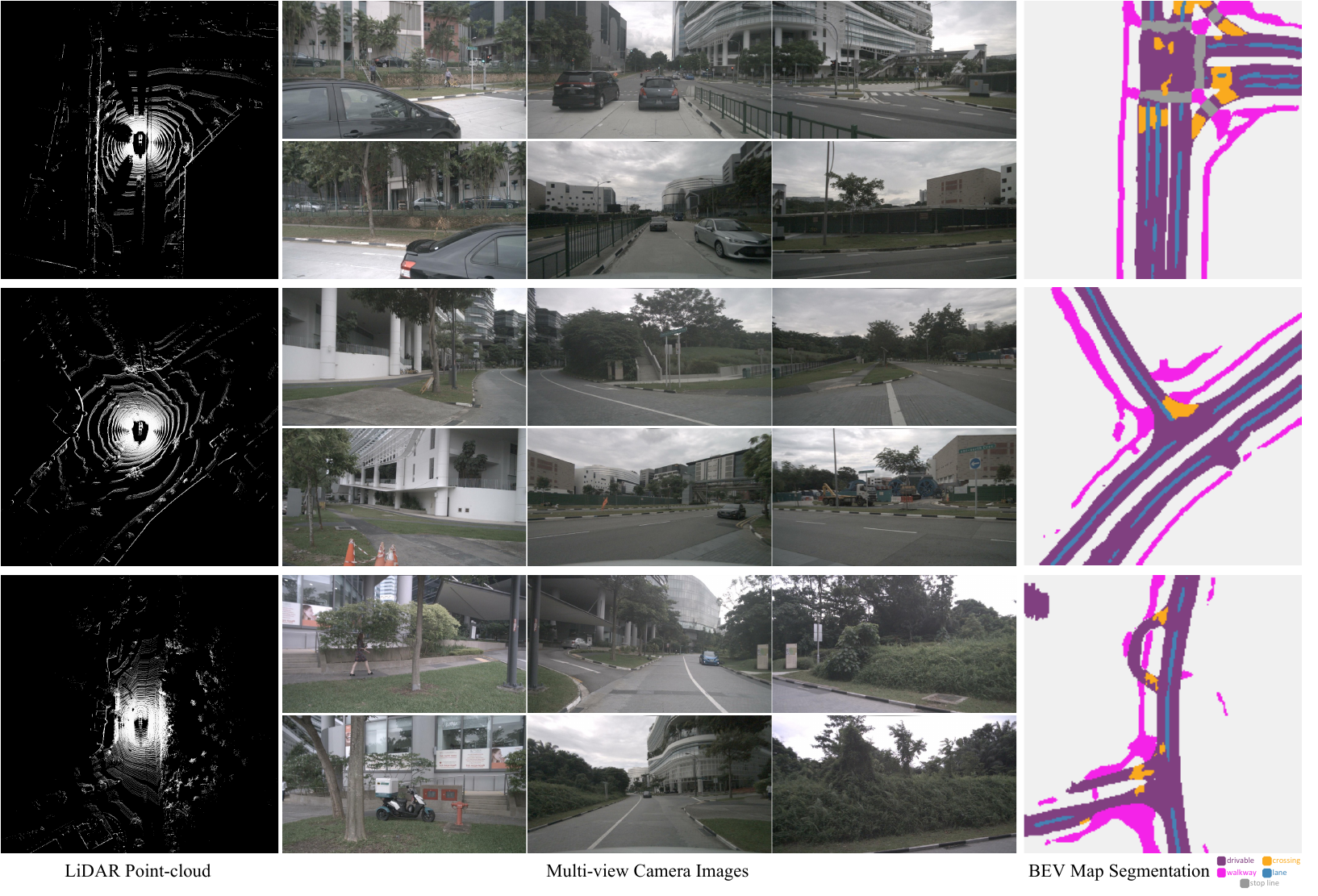}
  \caption{\textbf{Qualitative Results.} This figure presents the predicted BEV map segmentations across scenes with diverse levels of complexity. The leftmost column illustrates the lidar point-clouds. The middle columns display the six surrounding cameras, arranged from left to right as follows: front-left, front, front-right, back-left, back, and back-right. The rightmost column features the predicted BEV segmentations, with class labels including car park, walkway, lane divider, drivable area, stop line, and pedestrian crossing. The ego-vehicle is positioned at the centre of each map, facing upwards.}
  \label{fig:qualitative}
\end{figure*}

In Figure~\ref{fig:qualitative}, we present a selection of our qualitative results, showcasing scenes of varying complexity. The first column of each set displays the 2D projected lidar point-cloud, while the subsequent columns illustrates images from the surrounding multi-view cameras, encompassing front, back, front-right, back-right, front-left, and back-left perspectives. These are followed by the predicted BEV segmentation maps. 
For additional qualitative results, please refer to the supplementary video file.

\begin{figure*}
  \centering
  \includegraphics[width=\linewidth]{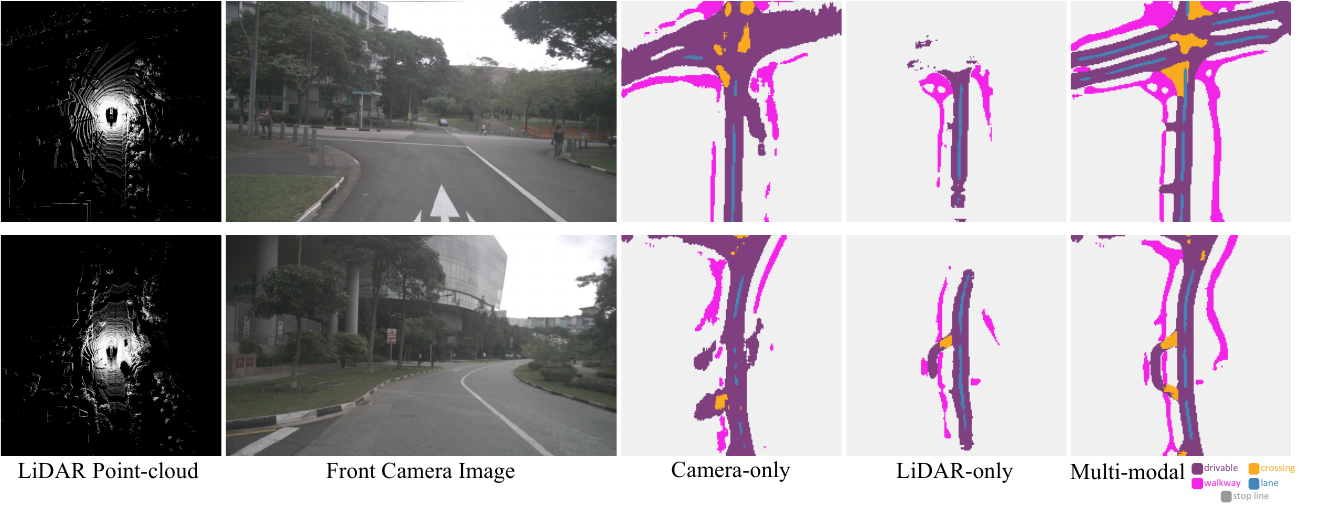}
  \caption{\textbf{Robustness Evaluation for Sensor Modality Dropout.} This figure assesses the robustness of our framework against sensor dropout, illustrating BEV segmentation predictions when various modalities are omitted. The first column displays the lidar point-cloud, and the second column shows the camera image (only the front camera is represented, with the other 5 camera views omitted for brevity). The leftmost predicted map results from using only the camera modality (excluding lidar), the middle map depicts predictions based on lidar-only, and the rightmost column illustrates predictions derived from fusing both modalities. The prediction maps confirm the framework's robustness to sensor dropout: the camera-only mode achieves broader coverage with less precise geometry, while the lidar-only mode provides enhanced geometric fidelity albeit over a more limited range. This robustness is crucial for the practical application of multi-modal fusion models in real-world autonomous driving scenarios.}
  \label{fig:robustness}
\end{figure*}

\subsection{Ablation Studies}

To validate our design decisions and analyse our framework's behaviour under diverse conditions, we investigated its performance across different parameter settings, including voxel size and depth discretisation resolution. In a separate study, we explored the impact of the modality selected for fine-tuning. After multi-modal self-supervised pretraining, we fine-tuned the model using ground-truth data, while freezing the backbone of one modality and fine-tuning the other. Table~\ref{tab:ablations_params} summarises the outcomes of these ablation studies, with bold figures denoting the performance in our optimally chosen configuration. As indicated in Table~\ref{tab:ablations_params}, fine-tuning with only camera data yields better performance compared to lidar-only, whilst a multi-modal approach delivers the best results.

\begin{table}
  \centering
  \caption{\textbf{Ablation Study of BEVPose.} This study examines the performance of our model under varying design parameters and with different combinations of input modalities for fine-tuning phase. The last column indicates mIoU. The bold number represents the performance in our final chosen setting.}
  \begin{tabular}{@{}lcc@{}}
    \toprule
                                   &    & mIoU \\
    \midrule
    \midrule
    \multirow{3}{*}{Fine-tuned Modality}      & Camera-only  & 61.0 \\
                                   & Lidar-only   & 53.2 \\
                                   & Camera+Lidar & \textbf{67.8} \\ 
    \midrule
    \midrule
    \multirow{2}{*}{Lidar Voxel Resolution}     & 0.2m     & 63.6 \\
                                                & 0.1m     & \textbf{67.8} \\ 
    \midrule
    \midrule
    \multirow{2}{*}{\#Depth Discrete Values ($L$)}    & 79     & 60.1 \\
                                                      & 158    & \textbf{67.8} \\ 
    \bottomrule
  \end{tabular}
  \label{tab:ablations_params}
\end{table}

\subsection{BEV Embeddings}
\begin{figure}
  \centering
  \includegraphics[width=\linewidth]{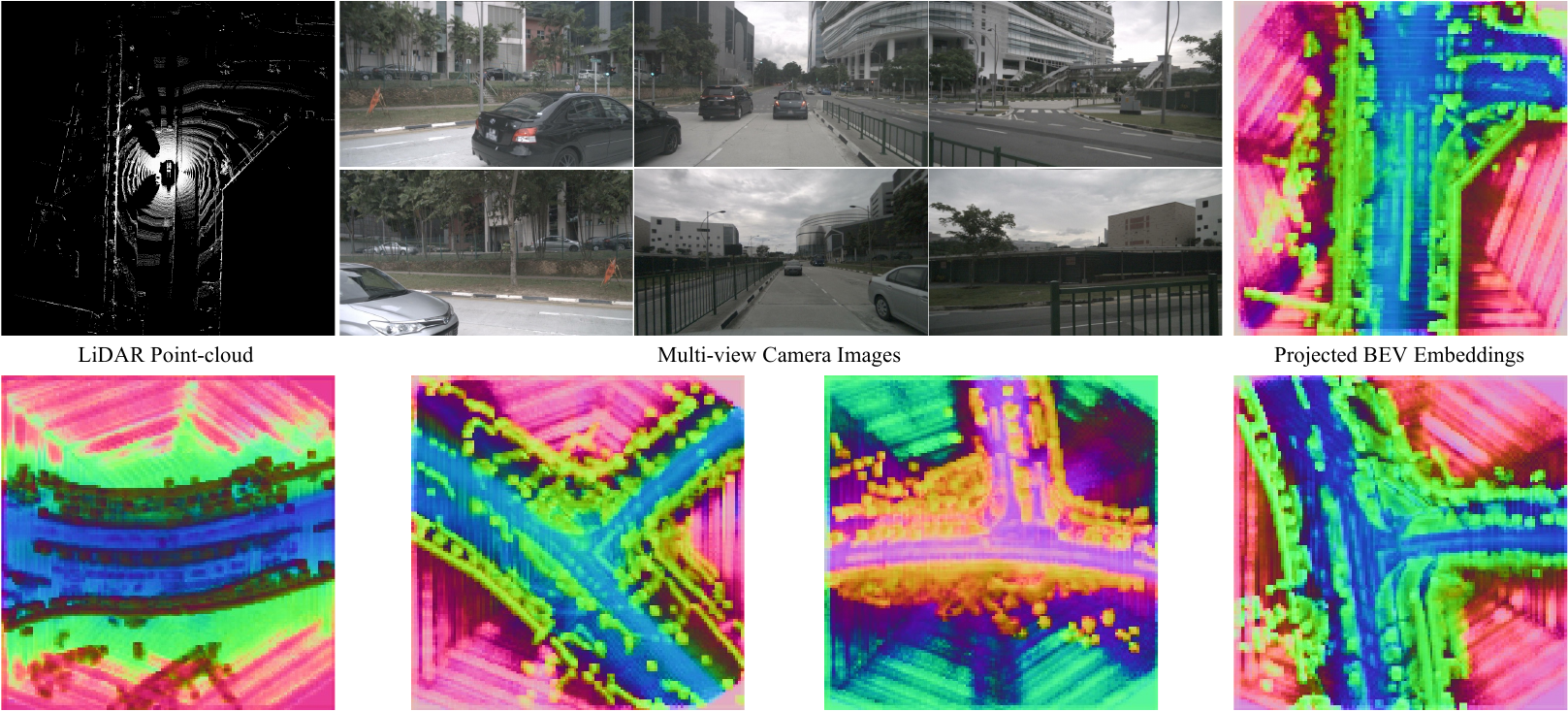}
  \caption{\textbf{BEV Embeddings.} An illustration of the self-supervised latent BEV 256-dimensional features projected into the RGB 3-space using PCA. The visualisation clearly demonstrates that these latent BEV representations capture high-level geometric and semantic information, such as roads, intersections, walkways, and other significant landmarks observable in camera frustums. The first row displays the lidar point-cloud, all 6 surrounding camera images, and their corresponding latent BEV embeddings. For brevity, only BEV features for various scenes are shown in the second row.}
  \label{fig:bev_embeddings}
\end{figure}

To conduct a qualitative assessment of the learned pose-supervised BEV embeddings, particularly their effectiveness in capturing high-level geometric and semantic information of the scene, we apply principal component analysis (PCA) to project the high-dimensional, fused latent BEV features into the RGB 3-space. As illustrated in Figure~\ref{fig:bev_embeddings}, this visualisation effectively unveils various semantic and geometric structures inherent in the BEV representation, including but not limited to roads, intersections, and the like. Furthermore, it distinctly highlights the camera BEV features emanating from the surrounding frustums.

\subsection{Robustness Evaluation}
Assessing the robustness to sensor dropout is a critical aspect of evaluating multi-modal fusion frameworks. Figure~\ref{fig:robustness} demonstrates the capability of our framework, which has undergone both pretraining and fine-tuning, to maintain performance integrity in the absence of one sensor modality during inference. The model exhibits a stable performance under conditions of sensor dropout. In scenarios utilising only the camera modality, the predicted BEV map spans wider ranges, albeit at a compromise to geometric precision. Conversely, the lidar-only mode, while constrained to nearer ranges, yields superior geometric accuracy.


%% file: sec/6_conclusion.tex
\section{Limitations}
\label{sec:supp_limitations}

BEVPose addresses a significant challenge in the recent trend of employing 2D BEV representations, specifically the pursuit of data-efficient method. It introduces a self-supervised, pose-guided, multi-modal BEV pretraining approach. However, this paper does not explore the integration of explicit multi-view camera constraints, such as stereo vision or depth sensing, into our pipeline — a direction we aim to pursue in future research. A practical scenario to consider is the application of stereo vision or RGB-D cameras in mobile robot navigation. Integrating explicit stereo-camera configurations into BEVPose could significantly expand its applicability for more diverse indoor and off-road environments.


\section{Conclusion}
\label{sec:conclusion}

BEVPose, our proposed framework, demonstrates an advancement in using self-supervised learning for fusing BEV representations from camera and lidar data. By leveraging sensor pose information, it effectively reduces reliance on extensive annotated data, particularly beneficial in diverse environments. The framework shows promising results in BEV map segmentation tasks, rivalling state-of-the-art methods with considerably less annotated data.

We aim to enhance BEVPose by incorporating temporal constraints using the trajectory data of autonomous agents. This could lead to more temporally coherent BEV representations. Additionally, we intend to explore the framework's application in varied environments such as off-road and indoor settings, and extend its use to other autonomous navigation tasks like 2D/3D object detection. Integrating other sensor modalities, such as depth sensors and radar, also presents an exciting avenue for further research.

%% file: root.bbl
\begin{thebibliography}{10}
\providecommand{\url}[1]{#1}
\csname url@rmstyle\endcsname
\providecommand{\newblock}{\relax}
\providecommand{\bibinfo}[2]{#2}
\providecommand\BIBentrySTDinterwordspacing{\spaceskip=0pt\relax}
\providecommand\BIBentryALTinterwordstretchfactor{4}
\providecommand\BIBentryALTinterwordspacing{\spaceskip=\fontdimen2\font plus
\BIBentryALTinterwordstretchfactor\fontdimen3\font minus
  \fontdimen4\font\relax}
\providecommand\BIBforeignlanguage[2]{{%
\expandafter\ifx\csname l@#1\endcsname\relax
\typeout{** WARNING: IEEEtran.bst: No hyphenation pattern has been}%
\typeout{** loaded for the language `#1'. Using the pattern for}%
\typeout{** the default language instead.}%
\else
\language=\csname l@#1\endcsname
\fi
#2}}

\bibitem{ORBSLAM3_TRO}
C.~Campos, R.~Elvira, J.~J. Gomez, J.~M.~M. Montiel, and J.~D. Tardos,
  ``{ORB-SLAM3}: An accurate open-source library for visual, visual-inertial
  and multi-map {SLAM},'' \emph{IEEE Transactions on Robotics}, vol.~37, no.~6,
  pp. 1874--1890, 2021.

\bibitem{orbslam}
R.~MurArtal, J.~M.~M. Montiel, and J.~D. Tardos, ``{ORB-SLAM}: a versatile and
  accurate monocular {SLAM} system,'' \emph{IEEE Transactions on Robotics},
  vol.~31, no.~5, pp. 1147--1163, 2015.

\bibitem{lsd-slam}
J.~Engel, T.~Sch{\"o}ps, and D.~Cremers, ``Lsd-slam: Large-scale direct
  monocular slam,'' in \emph{Computer Vision -- ECCV 2014}, D.~Fleet,
  T.~Pajdla, B.~Schiele, and T.~Tuytelaars, Eds.\hskip 1em plus 0.5em minus
  0.4em\relax Cham: Springer International Publishing, 2014, pp. 834--849.

\bibitem{schoenberger2016mvs}
J.~L. Sch\"{o}nberger, E.~Zheng, M.~Pollefeys, and J.-M. Frahm, ``Pixelwise
  view selection for unstructured multi-view stereo,'' in \emph{European
  Conference on Computer Vision (ECCV)}, 2016.

\bibitem{schoenberger2016sfm}
J.~L. Sch\"{o}nberger and J.-M. Frahm, ``Structure-from-motion revisited,'' in
  \emph{Conference on Computer Vision and Pattern Recognition (CVPR)}, 2016.

\bibitem{engel2016direct}
J.~Engel, V.~Koltun, and D.~Cremers, ``Direct sparse odometry,'' 2016.

\bibitem{legoloam2018}
T.~Shan and B.~Englot, ``Lego-loam: Lightweight and ground-optimized lidar
  odometry and mapping on variable terrain,'' in \emph{IEEE/RSJ International
  Conference on Intelligent Robots and Systems (IROS)}.\hskip 1em plus 0.5em
  minus 0.4em\relax IEEE, 2018, pp. 4758--4765.

\bibitem{wang2021}
H.~{Wang}, C.~{Wang}, C.~{Chen}, and L.~{Xie}, ``F-loam : Fast lidar odometry
  and mapping,'' in \emph{2021 IEEE/RSJ International Conference on Intelligent
  Robots and Systems (IROS)}, 2020.

\bibitem{bloesch2019codeslam}
M.~Bloesch, J.~Czarnowski, R.~Clark, S.~Leutenegger, and A.~J. Davison,
  ``Codeslam - learning a compact, optimisable representation for dense visual
  slam,'' 2019.

\bibitem{bylow2013real}
E.~Bylow, J.~Sturm, C.~Kerl, F.~Kahl, and D.~Cremers, ``Real-time camera
  tracking and 3d reconstruction using signed distance functions.'' in
  \emph{Robotics: Science and Systems}, vol.~2, 2013, p.~2.

\bibitem{wang2021nerf}
Z.~Wang, S.~Wu, W.~Xie, M.~Chen, and V.~A. Prisacariu, ``Nerf--: Neural
  radiance fields without known camera parameters,'' \emph{arXiv preprint
  arXiv:2102.07064}, 2021.

\bibitem{pumarola2021d}
A.~Pumarola, E.~Corona, G.~Pons-Moll, and F.~Moreno-Noguer, ``D-nerf: Neural
  radiance fields for dynamic scenes,'' in \emph{Proceedings of the IEEE/CVF
  Conference on Computer Vision and Pattern Recognition}, 2021, pp.
  10\,318--10\,327.

\bibitem{yang2019cubeslam}
S.~Yang and S.~Scherer, ``Cubeslam: Monocular 3-d object slam,'' \emph{IEEE
  Transactions on Robotics}, vol.~35, no.~4, pp. 925--938, 2019.

\bibitem{taguchi2013point}
Y.~Taguchi, Y.-D. Jian, S.~Ramalingam, and C.~Feng, ``Point-plane slam for
  hand-held 3d sensors,'' in \emph{2013 IEEE international conference on
  robotics and automation}.\hskip 1em plus 0.5em minus 0.4em\relax IEEE, 2013,
  pp. 5182--5189.

\bibitem{hosseinzadeh2019real}
M.~Hosseinzadeh, K.~Li, Y.~Latif, and I.~Reid, ``Real-time monocular
  object-model aware sparse slam,'' in \emph{2019 International Conference on
  Robotics and Automation (ICRA)}.\hskip 1em plus 0.5em minus 0.4em\relax IEEE,
  2019, pp. 7123--7129.

\bibitem{hosseinzadeh2019structure}
M.~Hosseinzadeh, Y.~Latif, T.~Pham, N.~Suenderhauf, and I.~Reid, ``Structure
  aware slam using quadrics and planes,'' in \emph{Computer Vision--ACCV 2018:
  14th Asian Conference on Computer Vision, Perth, Australia, December 2--6,
  2018, Revised Selected Papers, Part III 14}.\hskip 1em plus 0.5em minus
  0.4em\relax Springer, 2019, pp. 410--426.

\bibitem{huang2021bevdet}
J.~Huang, G.~Huang, Z.~Zhu, Y.~Yun, and D.~Du, ``Bevdet: High-performance
  multi-camera 3d object detection in bird-eye-view,'' \emph{arXiv preprint
  arXiv:2112.11790}, 2021.

\bibitem{huang2022bevdet4d}
J.~Huang and G.~Huang, ``Bevdet4d: Exploit temporal cues in multi-camera 3d
  object detection,'' \emph{arXiv preprint arXiv:2203.17054}, 2022.

\bibitem{bevfusion2022}
Z.~Liu, H.~Tang, A.~Amini, X.~Yang, H.~Mao, D.~Rus, and S.~Han, ``Bevfusion:
  Multi-task multi-sensor fusion with unified bird's-eye view representation,''
  in \emph{IEEE International Conference on Robotics and Automation (ICRA)},
  2023.

\bibitem{liang2022bevfusion}
T.~Liang, H.~Xie, K.~Yu, Z.~Xia, Z.~Lin, Y.~Wang, T.~Tang, B.~Wang, and
  Z.~Tang, ``{BEVFusion: A Simple and Robust LiDAR-Camera Fusion Framework},''
  in \emph{Neural Information Processing Systems (NeurIPS)}, 2022.

\bibitem{sarlin2023orienternet}
P.-E. Sarlin, D.~DeTone, T.-Y. Yang, A.~Avetisyan, J.~Straub, T.~Malisiewicz,
  S.~R. Bulo, R.~Newcombe, P.~Kontschieder, and V.~Balntas, ``{OrienterNet:
  Visual Localization in 2D Public Maps with Neural Matching},'' in
  \emph{CVPR}, 2023.

\bibitem{sarlin2023snap}
P.-E. Sarlin, E.~Trulls, M.~Pollefeys, J.~Hosang, and S.~Lynen, ``{SNAP:
  Self-Supervised Neural Maps for Visual Positioning and Semantic
  Understanding},'' in \emph{NeurIPS}, 2023.

\bibitem{yin2020center}
T.~Yin, X.~Zhou, and P.~Kr{\"a}henb{\"u}hl, ``Center-based 3d object detection
  and tracking,'' \emph{arXiv preprint arXiv:2006.11275}, 2020.

\bibitem{huang2023dal}
J.~Huang, Y.~Ye, Z.~Liang, Y.~Shan, and G.~Huang, ``Detecting as labeling:
  Rethinking lidar-camera fusion in 3d object detection,'' \emph{arXiv preprint
  arXiv:2311.07152}, 2023.

\bibitem{huang2022bevpoolv2}
J.~Huang and G.~Huang, ``Bevpoolv2: A cutting-edge implementation of bevdet
  toward deployment,'' \emph{arXiv preprint arXiv:2211.17111}, 2022.

\bibitem{tian2019fcos}
Z.~Tian, C.~Shen, H.~Chen, and T.~He, ``Fcos: Fully convolutional one-stage
  object detection,'' in \emph{Proceedings of the IEEE/CVF International
  Conference on Computer Vision}, 2019, pp. 9627--9636.

\bibitem{carion2020end}
N.~Carion \emph{et~al.}, ``End-to-end object detection with transformers,''
  \emph{arXiv preprint arXiv:2005.12872}, 2020.

\bibitem{chen2017multi}
X.~Chen, H.~Ma, J.~Wan, B.~Li, and T.~Xia, ``Multi-view 3d object detection
  network for autonomous driving,'' in \emph{Proceedings of the IEEE Conference
  on Computer Vision and Pattern Recognition}, 2017, pp. 1907--1915.

\bibitem{qi2018frustum}
C.~R. Qi, W.~Liu, C.~Wu, H.~Su, and L.~J. Guibas, ``Frustum pointnets for 3d
  object detection from rgb-d data,'' in \emph{Proceedings of the IEEE
  conference on computer vision and pattern recognition}, 2018, pp. 918--927.

\bibitem{wang2021end}
J.~Wang \emph{et~al.}, ``End-to-end multi-modal multi-task vehicle control for
  self-driving cars with visual perceptions,'' in \emph{Proceedings of the
  IEEE/CVF International Conference on Computer Vision}, 2021, pp. 8858--8867.

\bibitem{reading2021categorical}
C.~Reading, A.~Harakeh, and S.~L. Waslander, ``Categorical depth distribution
  network for monocular 3d object detection,'' \emph{arXiv preprint
  arXiv:2103.01100}, 2021.

\bibitem{philion2020lift}
J.~Philion and S.~Fidler, ``Lift, splat, shoot: Encoding images from arbitrary
  camera rigs by implicitly unprojecting to 3d,'' \emph{arXiv preprint
  arXiv:2008.05711}, 2020.

\bibitem{you2019pseudo}
Y.~You, Y.~Wang, W.-L. Chao, D.~Garg, G.~Pleiss, B.~Hariharan, M.~Campbell, and
  K.~Q. Weinberger, ``Pseudo-lidar++: Accurate depth for 3d object detection in
  autonomous driving,'' \emph{arXiv preprint arXiv:1906.06310}, 2019.

\bibitem{li2023voxformer}
Y.~Li, Z.~Yu, C.~Choy, C.~Xiao, J.~M. Alvarez, S.~Fidler, C.~Feng, and
  A.~Anandkumar, ``Voxformer: Sparse voxel transformer for camera-based 3d
  semantic scene completion,'' in \emph{Proceedings of the IEEE/CVF conference
  on computer vision and pattern recognition}, 2023, pp. 9087--9098.

\bibitem{saha2022translating}
A.~Saha, O.~Mendez, C.~Russell, and R.~Bowden, ``Translating images into
  maps,'' in \emph{2022 International conference on robotics and automation
  (ICRA)}.\hskip 1em plus 0.5em minus 0.4em\relax IEEE, 2022, pp. 9200--9206.

\bibitem{zhou2022cross}
B.~Zhou and P.~Kr{\"a}henb{\"u}hl, ``Cross-view transformers for real-time
  map-view semantic segmentation,'' in \emph{CVPR}, 2022.

\bibitem{li2022bevformer}
Z.~Li, W.~Wang, H.~Li, E.~Xie, C.~Sima, T.~Lu, Y.~Qiao, and J.~Dai,
  ``Bevformer: Learning bird’s-eye-view representation from multi-camera
  images via spatiotemporal transformers,'' in \emph{European conference on
  computer vision}.\hskip 1em plus 0.5em minus 0.4em\relax Springer, 2022, pp.
  1--18.

\bibitem{peng2023bevsegformer}
L.~Peng, Z.~Chen, Z.~Fu, P.~Liang, and E.~Cheng, ``Bevsegformer: Bird's eye
  view semantic segmentation from arbitrary camera rigs,'' in \emph{Proceedings
  of the IEEE/CVF Winter Conference on Applications of Computer Vision}, 2023,
  pp. 5935--5943.

\bibitem{zhu2020deformable}
X.~Zhu, W.~Su, L.~Lu, B.~Li, X.~Wang, and J.~Dai, ``Deformable detr: Deformable
  transformers for end-to-end object detection,'' in \emph{International
  Conference on Learning Representations}, 2020.

\bibitem{liu2022petr}
Y.~Liu, T.~Wang, X.~Zhang, and J.~Sun, ``Petr: Position embedding
  transformation for multi-view 3d object detection,'' \emph{arXiv preprint
  arXiv:2203.05625}, 2022.

\bibitem{liu2022petrv2}
Y.~Liu, J.~Yan, F.~Jia, S.~Li, Q.~Gao, T.~Wang, X.~Zhang, and J.~Sun, ``Petrv2:
  A unified framework for 3d perception from multi-camera images,'' \emph{arXiv
  preprint arXiv:2206.01256}, 2022.

\bibitem{zhang2022beverse}
Y.~Zhang, Z.~Zhu, W.~Zheng, J.~Huang, G.~Huang, J.~Zhou, and J.~Lu, ``Beverse:
  Unified perception and prediction in birds-eye-view for vision-centric
  autonomous driving,'' \emph{arXiv preprint arXiv:2205.09743}, 2022.

\bibitem{vandenoord2018representation}
A.~van~den Oord, Y.~Li, and O.~Vinyals, ``Representation learning with
  contrastive predictive coding,'' \emph{arXiv preprint arXiv:1807.03748},
  2018.

\bibitem{he2020momentum}
K.~He, H.~Fan, Y.~Wu, S.~Xie, and R.~Girshick, ``Momentum contrast for
  unsupervised visual representation learning,'' in \emph{Proceedings of the
  IEEE/CVF Conference on Computer Vision and Pattern Recognition}, 2020, pp.
  9726--9735.

\bibitem{caron2021emerging}
M.~Caron, H.~Touvron, I.~Misra, H.~J{\'e}gou, J.~Mairal, P.~Bojanowski, and
  A.~Joulin, ``Emerging properties in self-supervised vision transformers,'' in
  \emph{Proceedings of the IEEE/CVF International Conference on Computer
  Vision}, 2021, pp. 9650--9660.

\bibitem{pinheiro2020unsupervised}
P.~O. Pinheiro, A.~Almahairi, R.~Benmalek, F.~Golemo, and A.~C. Courville,
  ``Unsupervised learning of dense visual representations,'' in \emph{Advances
  in Neural Information Processing Systems}, vol.~33, 2020, pp. 7048--7060.

\bibitem{he2022masked}
K.~He, X.~Chen, S.~Xie, Y.~Li, P.~Doll{\'a}r, and R.~Girshick, ``Masked
  autoencoders are scalable vision learners,'' in \emph{Proceedings of the
  IEEE/CVF Conference on Computer Vision and Pattern Recognition}, 2022, pp.
  16\,000--16\,009.

\bibitem{larsson2019fine}
M.~Larsson, E.~Stenborg, C.~Toft, L.~Hammarstrand, T.~Sattler, and F.~Kahl,
  ``Fine-grained segmentation networks: Self-supervised segmentation for
  improved long-term visual localization,'' in \emph{Proceedings of the
  IEEE/CVF International Conference on Computer Vision}, 2019, pp. 9915--9924.

\bibitem{mildenhall2020nerf}
B.~Mildenhall, P.~P. Srinivasan, M.~Tancik, J.~T. Barron, R.~Ramamoorthi, and
  R.~Ng, ``Nerf: Representing scenes as neural radiance fields for view
  synthesis,'' in \emph{European Conference on Computer Vision}.\hskip 1em plus
  0.5em minus 0.4em\relax Springer, 2020, pp. 405--421.

\bibitem{sharma2022seeing}
P.~Sharma, A.~Tewari, Y.~Du, S.~Zakharov, R.~Ambrus, A.~Gaidon, W.~T. Freeman,
  F.~Durand, J.~B. Tenenbaum, and V.~Sitzmann, ``Seeing 3d objects in a single
  image via self-supervised static-dynamic disentanglement,'' in \emph{Advances
  in Neural Information Processing Systems}, 2022.

\bibitem{lal2021coconets}
S.~Lal, M.~Prabhudesai, I.~Mediratta, A.~W. Harley, and K.~Fragkiadaki,
  ``Coconets: Continuous contrastive 3d scene representations,'' in
  \emph{Proceedings of the IEEE/CVF Conference on Computer Vision and Pattern
  Recognition}, 2021, pp. 16\,995--17\,004.

\bibitem{bai2022transfusion}
X.~Bai, Z.~Hu, X.~Zhu, Q.~Huang, Y.~Chen, H.~Fu, and C.-L. Tai, ``Transfusion:
  Robust lidar-camera fusion for 3d object detection with transformers,'' in
  \emph{Proceedings of the IEEE/CVF conference on computer vision and pattern
  recognition}, 2022, pp. 1090--1099.

\bibitem{he2016deep}
K.~He, X.~Zhang, S.~Ren, and J.~Sun, ``Deep residual learning for image
  recognition,'' in \emph{Proceedings of the IEEE conference on computer vision
  and pattern recognition}, 2016, pp. 770--778.

\bibitem{liu2021swin}
Z.~Liu, Y.~Lin, Y.~Cao, H.~Hu, Y.~Wei, Z.~Zhang, S.~Lin, and B.~Guo, ``Swin
  transformer: Hierarchical vision transformer using shifted windows,'' in
  \emph{Proceedings of the IEEE/CVF international conference on computer
  vision}, 2021, pp. 10\,012--10\,022.

\bibitem{lin2017feature}
T.-Y. Lin, P.~Doll{\'a}r, R.~Girshick, K.~He, B.~Hariharan, and S.~Belongie,
  ``Feature pyramid networks for object detection,'' in \emph{Proceedings of
  the IEEE conference on computer vision and pattern recognition}, 2017, pp.
  2117--2125.

\bibitem{zhou2018voxelnet}
Y.~Zhou and O.~Tuzel, ``Voxelnet: End-to-end learning for point cloud based 3d
  object detection,'' in \emph{Proceedings of the IEEE conference on computer
  vision and pattern recognition}, 2018, pp. 4490--4499.

\bibitem{loshchilov2017decoupled}
I.~Loshchilov and F.~Hutter, ``Decoupled weight decay regularization,''
  \emph{arXiv preprint arXiv:1711.05101}, 2017.

\bibitem{caesar2020nuscenes}
H.~Caesar, V.~Bankiti, A.~H. Lang, S.~Vora, V.~E. Liong, Q.~Xu, A.~Krishnan,
  Y.~Pan, G.~Baldan, and O.~Beijbom, ``nuscenes: A multimodal dataset for
  autonomous driving,'' in \emph{Proceedings of the IEEE/CVF conference on
  computer vision and pattern recognition}, 2020, pp. 11\,621--11\,631.

\bibitem{lin2017focal}
T.-Y. Lin, P.~Goyal, R.~Girshick, K.~He, and P.~Doll{\'a}r, ``Focal loss for
  dense object detection,'' in \emph{Proceedings of the IEEE international
  conference on computer vision}, 2017, pp. 2980--2988.

\bibitem{roddick2018orthographic}
T.~Roddick, A.~Kendall, and R.~Cipolla, ``Orthographic feature transform for
  monocular 3d object detection,'' \emph{arXiv preprint arXiv:1811.08188},
  2018.

\bibitem{xie2022m}
E.~Xie, Z.~Yu, D.~Zhou, J.~Philion, A.~Anandkumar, S.~Fidler, P.~Luo, and J.~M.
  Alvarez, ``M$^2$bev: Multi-camera joint 3d detection and segmentation with
  unified birds-eye view representation,'' \emph{arXiv preprint
  arXiv:2204.05088}, 2022.

\bibitem{yin2021center}
T.~Yin, X.~Zhou, and P.~Krahenbuhl, ``Center-based 3d object detection and
  tracking,'' in \emph{Proceedings of the IEEE/CVF conference on computer
  vision and pattern recognition}, 2021, pp. 11\,784--11\,793.

\bibitem{lang2019pointpillars}
A.~H. Lang, S.~Vora, H.~Caesar, L.~Zhou, J.~Yang, and O.~Beijbom,
  ``Pointpillars: Fast encoders for object detection from point clouds,'' in
  \emph{Proceedings of the IEEE/CVF conference on computer vision and pattern
  recognition}, 2019, pp. 12\,697--12\,705.

\bibitem{yin2021multimodal}
T.~Yin, X.~Zhou, and P.~Kr{\"a}henb{\"u}hl, ``Multimodal virtual point 3d
  detection,'' \emph{Advances in Neural Information Processing Systems},
  vol.~34, pp. 16\,494--16\,507, 2021.

\bibitem{vora2020pointpainting}
S.~Vora, A.~H. Lang, B.~Helou, and O.~Beijbom, ``Pointpainting: Sequential
  fusion for 3d object detection,'' in \emph{Proceedings of the IEEE/CVF
  conference on computer vision and pattern recognition}, 2020, pp. 4604--4612.

\end{thebibliography}
